\documentclass[conference]{IEEEtran}
\IEEEoverridecommandlockouts
\usepackage{cite}
\usepackage{amsmath,amssymb,amsfonts}
\usepackage{graphicx}
\usepackage[]{hyperref}
\usepackage{textcomp}
\usepackage{xcolor}
\usepackage{algorithm}
\usepackage{algpseudocode}

\def\BibTeX{{\rm B\kern-.05em{\sc i\kern-.025em b}\kern-.08em
    T\kern-.1667em\lower.7ex\hbox{E}\kern-.125emX}}

\begin{document}
\title{W-Transformers : A Wavelet-based Transformer Framework for Univariate Time Series Forecasting}


 \author{\IEEEauthorblockN{Lena Sasal}
 \IEEEauthorblockA{\textit{Sorbonne Center for Artificial Intelligence} \\
 \textit{Sorbonne University Abu Dhabi}\\
 Abu Dhabi, UAE\\
 Lena.Sasal@sorbonne.ae}
 \and
 \IEEEauthorblockN{Tanujit Chakraborty}
 \IEEEauthorblockA{\textit{Dept. of Science and Engineering} \\
 \textit{Sorbonne University Abu Dhabi}\\
 Abu Dhabi, UAE\\
 tanujit.chakraborty@sorbonne.ae}
 \and
 \IEEEauthorblockN{Abdenour Hadid}
 \IEEEauthorblockA{\textit{Sorbonne Center for Artificial Intelligence} \\
 \textit{Sorbonne University Abu Dhabi}\\
 Abu Dhabi, UAE\\
 Abdenour.Hadid@sorbonne.ae}
 }

\maketitle

\begin{abstract}
Deep learning utilizing transformers has recently achieved a lot of success in many vital areas such as natural language processing, computer vision, anomaly detection, and recommendation systems, among many others. Among several merits of transformers, the ability to capture long-range temporal dependencies and interactions is desirable for time series forecasting, leading to its progress in various time series applications. In this paper, we build a transformer model for non-stationary time series. The problem is challenging yet crucially important. We present a novel framework for univariate time series representation learning based on the wavelet-based transformer encoder architecture and call it W-Transformer. The proposed W-Transformers utilize a maximal overlap discrete wavelet transformation (MODWT) to the time series data and build local transformers on the decomposed datasets to vividly capture the nonstationarity and long-range nonlinear dependencies in the time series. Evaluating our framework on several publicly available benchmark time series datasets from various domains and with diverse characteristics, we demonstrate that it performs, on average, significantly better than the baseline forecasters for short-term and long-term forecasting, even for datasets that consist of only a few hundred training samples. 
\end{abstract}


\section{Introduction}
Forecasting the future movement and value of time series is a key component of formulating effective strategies in most business, industrial fields, and medical domains, among many others \cite{hyndman2018forecasting, hyndman2008forecasting}. Specific applications include forecasting epidemiological cases (including Covid-19) \cite{chakraborty2020real, panja2022epicasting}, stock price prediction \cite{Mehtab_2020}, weather forecasts \cite{REN2021100178}, web traffic forecasting \cite{Casado-Vara2021WebTraffic} and others applied fields \cite{vautard2001earth,qin2019wind}. Over the last decades, many techniques have evolved in time series forecasting literature. Statistics models (such as autoregressive integrated moving average (ARIMA) \cite{box2015time}, state-space exponential smoothing (ETS) \cite{hyndman2008forecasting}, and self-excited threshold autoregressive (SETAR) \cite{tong2009threshold} models) have been intensively used in several applied domains due to their easy interpretability. However, they tend to fail when highly complex situations arise (for example, nonlinear and nonstationary time series) and require expertise in manually selecting trends, seasonality, and other components. On the other hand, machine learning methods such as LightGBM \cite{ke2017GBM} and autoregressive neural networks (ARNN) \cite{faraway1998time} can leverage the ground truth data to learn the trends and patterns in an automated manner while sometimes failing when there is long-term dependence on the observed time series data.  

Current progress in deep learning has brought us deep autoregressive (DeepAR) \cite{salinas2020deepar} models, temporal convolutional networks (TCN) \cite{bai2018TCN} and many others \cite{petropoulos2022forecasting, lim2021time} which can deal with nonlinear and non stationary time series. Time series data have also been modeled using Recurrent Neural Network (RNN) \cite{Hewamalage2019RNN} in an autoregressive way that can tackle the irregularity of the time series. However, due to the gradient vanishing and exploding problem, RNNs are challenging to train. The problems continue to exist despite the introduction of numerous versions, such as long short-term memory (LSTM) \cite{Hochreiter1997LSTM} and gated recurrent unit (GRU) \cite{cho2018GRconv} that often outperform standard deep models. However, practical forecasting applications frequently contain both long and short-term recurrent patterns, and RNN-based methods sometimes struggle to capture long-term dependencies. 

The introduction of Transformers in deep learning \cite{vaswani2017attention} has brought great interests for researchers in natural language processing \cite{wolf2020transformers}, computer vision \cite{dosovitskiy2020image}, and other disciplines \cite{wen2022transformers}. Transformers have shown great success in modeling and extrapolating the long-range dependencies and interactions in temporal data; hence they are pertinent in time series forecasting \cite{godahewa2021monash}. Among various applications of Transformers for time series forecasting, some seminal papers are: probabilistic Transformer for multivariate time series representation learning \cite{zerveas2021transformer, tang2021probabilistic}, deep transformer models for epidemic forecasting \cite{wu2020deep} and many others \cite{wen2022transformers}. The multi-headed attention mechanism of Transformers makes them particularly suitable for time series data analysis: (a) they concurrently represent each input sequence element by considering its context, (b) while multiple attention heads can consider different representation subspaces, i.e., multiple aspects of relevance between input elements -- for time series, this, for example, may correspond to multiple periodicities in the signal \cite{zerveas2021transformer}. Other applications of Transformers include spatio-temporal forecasting in which, introducing a temporal Transformer block to capture temporal dependencies, it is also essential to design a spatial Transformer block \cite{wen2022transformers}. Transformers are also proven effective in various time series classification tasks due to their superior capability in capturing long-term dependency \cite{feng2022multi}. Compared to RNN and other neural network structures, the advantage of Transformers is that they easily enable parallel processing and reduce computational time since they do not involve recurrent behavior. A remaining challenge within Transformers is how to effectively model long-range and short-range temporal dependency and capture nonstationarity and nonlinearity simultaneously for forecasting time series data. 

So far, many sophisticated decomposition techniques have shown strong competitiveness in time series pre-processing tasks, such as Fourier transformation \cite{sneddon1995fourier}, fast Fourier transformation \cite{cochran1967fast}, and Wavelet decomposition \cite{percival2000wavelet} to cite a few. These methods of decomposing the complex time series into simpler parts have often led to satisfactory results. Among them, Wavelet decomposition, a mathematical tool that can reveal information within the signals in both the time and scale (frequency) domains, overcomes the basic drawback of Fourier analysis \cite{chakraborty2020real}. Most of the previous applications of wavelets have used discrete wavelet transform (DWT) and during the task of forecasting, it has been combined with statistical or machine learning algorithm \cite{liu2013forecasting, boto2010wavelet}. The limitation of DWT on the signal length made us use the Maximal Overlap discrete transform (MODWT) algorithm which counters this limitation but has the same abilities \cite{percival1997analysis}. In fact, MODWT preserves the number of coefficients of the original time series for all the levels in the decomposition. Wavelet-ARIMA (WARIMA) \cite{aminghafari2007forecasting} and Wavelet neural network \cite{panja2022epicasting} are some of the hybrid algorithms that combine MODWT algorithm and statistical or machine learning models. Those hybrid approaches have been used in several practical applications, for example, epidemics \cite{panja2022epicasting}, hydrology \cite{sang2013review}, traffic volumes \cite{boto2010wavelet}, and wind speed \cite{liu2013forecasting} forecasting problems.  

Motivated by the above observations, we propose a novel framework combining the MODWT algorithm with Transformers to effectively model the long-range and short-range temporal dependency and capture nonstationarity and nonlinearity simultaneously of the time series. Incorporating wavelets within Transformers will enable the model to capture all the irregularities of the time series by having a more detailed decomposition of low and high-frequency coefficients to separate the signal from the noise. Furthermore, scaling the wavelet used on the signal will reveal more about the frequency or the time. In fact, stretching the wavelet will give more information about the frequency while losing some time localization. Similarly, squeezing the wavelet will result in losing a part of the frequency domain while better preserving time location. Given those observations, our proposed W-Transformers is expected to generate more accurate and stable out-of-sample forecasts for long-term horizons for nonstationary time series data. 
Our main contributions to this paper can be described as follows : 
\begin{itemize}
    \item We propose a novel model (namely  W-Transformers) for time series forecasting elegantly combining wavelet decomposition and Transformers in the ensemble approach. 
    \item The proposed W-Transformers possess the capacity to handle long-range dependencies and nonlinear behavior (with the help of Transformers) to capture the nonstationarity of the time series (with the use of Wavelet decomposition) in its ensemble framework. Thus, it is expected to be more reliable and accurate than conventional transformers and other deep learning architectures. 
    \item We perform extensive experiments with multiple univariate time series datasets, demonstrating superior performance compared with the current state-of-the-art forecast methods for short-term and long-term predictions.
\end{itemize}

\section{Mathematical Preliminaries}
This section discusses the mathematical foundation of the wavelet decomposition to be used in our proposed W-Transformers model. For more details, one may refer to \cite{percival2000wavelet}.  

\subsection{Wavelet Decomposition}
Wavelet is a wave-like oscillation localized in time. It can capture the time and the frequency of a signal, contrary to Fourier Transform, which only captures the frequency of the signal. There are two different kinds of wavelets: (a) Continuous wavelet transforms (CWT), which applies an infinite number of wavelet (i.e., every possible wavelet) and (b) Discrete wavelet transform (DWT), where a finite number of wavelets in specific time and location are applied. In our work, we focus on DWT that represents a signal using an orthonormal basis representation, widely used in hydrology \cite{sang2013review}, epidemics \cite{panja2022epicasting}, and geophysics \cite{grinsted2004application} to name a few. This decomposition will give us a set of time series where each time series has a coefficient describing the evolution in time of the signal in a specific frequency band. The DWT uses a discrete dyadic (octave) grid for scale parameter $i$ and shift parameter $k$, and the equation for forward DWT is as follows \cite{percival2000wavelet}:
$$
c(i, k)=\sum_{t} f(t) \Psi_{i, k}^{*}(t), \text { where } \Psi_{i, k}^{*}(t)=2^{\frac{i}{2}} \Psi\left(2^{i} t-k\right).
$$
The mathematical equation for inverse DWT is as follows:
$$
f(t)=\sum_{k} \sum_{i} c(i, k) \Psi_{i, k}(t),
$$
where $f(\cdot)$ is a function and $\Psi$ the mother wavelet \cite{percival2000wavelet}.  

\subsection{MODWT algorithm}\label{modwt}
Since DWT restricts the sample size to be exactly a power of $2$, a modified version of the DWT, namely maximal overlapping discrete wavelet transformations (MODWT), is used for decomposing arbitrary time series \cite{percival1997analysis}. Both MODWT and DWT can handle multi-resolution analysis, which is a scaled-based additive decomposition, but in the case of the MODWT, it also has the capacity to catch the circular shift of the signal. The shift-invariance property of MODWT is an added advantage over DWT in its time series applications. Below, we define the detail coefficients $(\Tilde{s}_{j,l})$ and scaling coefficients $(\Tilde{s}_{j,l})$ that are produced by an MODWT algorithm with Haar filter (most commonly used wavelet filter in time series literature) \cite{zhu2014modwt}. 
$$ \Tilde{d}_{j,l} = {d_{j,l}}/2^{\frac{j}{2}} \; \; \text{and} \; \; {\Tilde{s}_{j,l}} = {s_{j,l}}/{2^{\frac{j}{2}}},$$
where $d_{j,l}$ and $s_{j,l}$ are wavelet and scaling filters of DWT algorithm. The coefficient of level $j$ $(l = 0,1,\ldots,L-1)$ then convolve the original time series $\{Y_t: t=0,1,\ldots,N-1\}$ and the MODWT pyramid algorithm generates the MODWT detail coefficients $(\Tilde{D}_{j,t})$ and scaling coefficients $(\Tilde{S}_{j,t})$ as follows \cite{percival2000wavelet}:
$$\Tilde{D}_{j,t}=\sum_{l=0}^{L_{j}-1}\Tilde{d}_{j,l}Y_{(t-l) mod N}
\; \; \text{and} \; \; 
\Tilde{S}_{j,t}=\sum_{l=0}^{L_{j}-1}\Tilde{s}_{j,l}Y_{(t-l) mod N},$$
where $L_{j} = (2^{j}-1)(L-1)+1$. All the levels and the smooth series will have the same length as the original one.

\section{Proposed Method: W-Transformers}\label{proposed}
\begin{figure}
    \centering
       \includegraphics[width=0.8\linewidth]{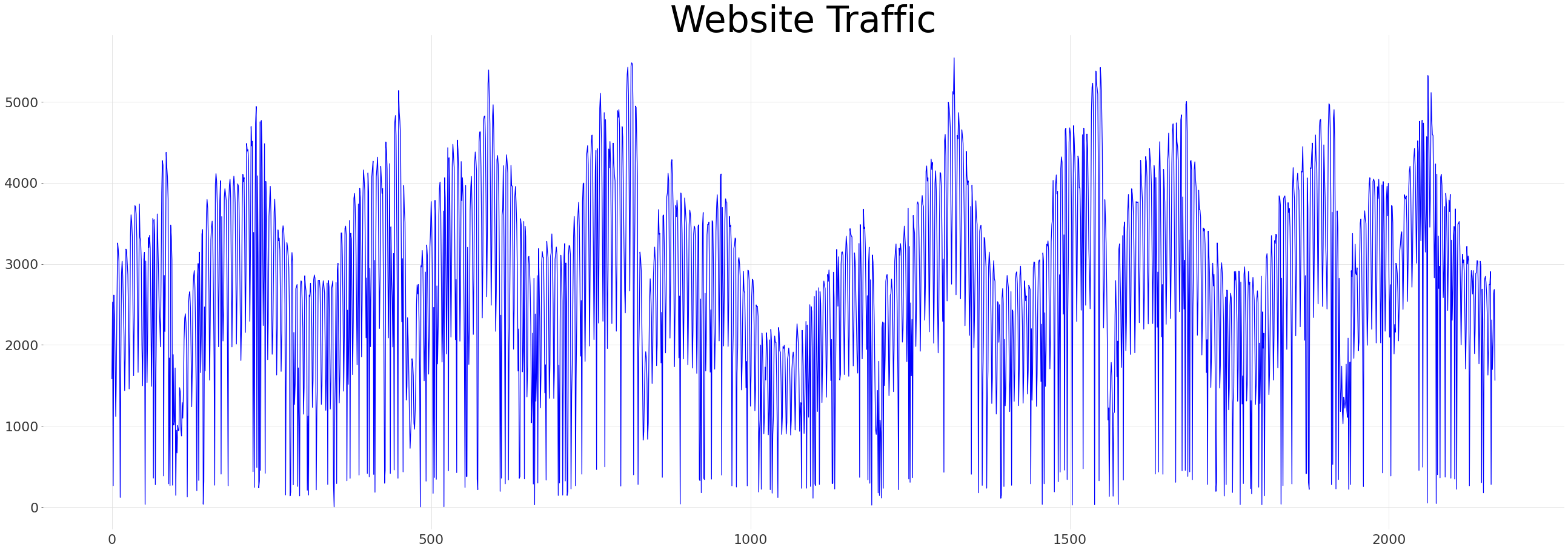}
       \includegraphics[width=0.8\linewidth]{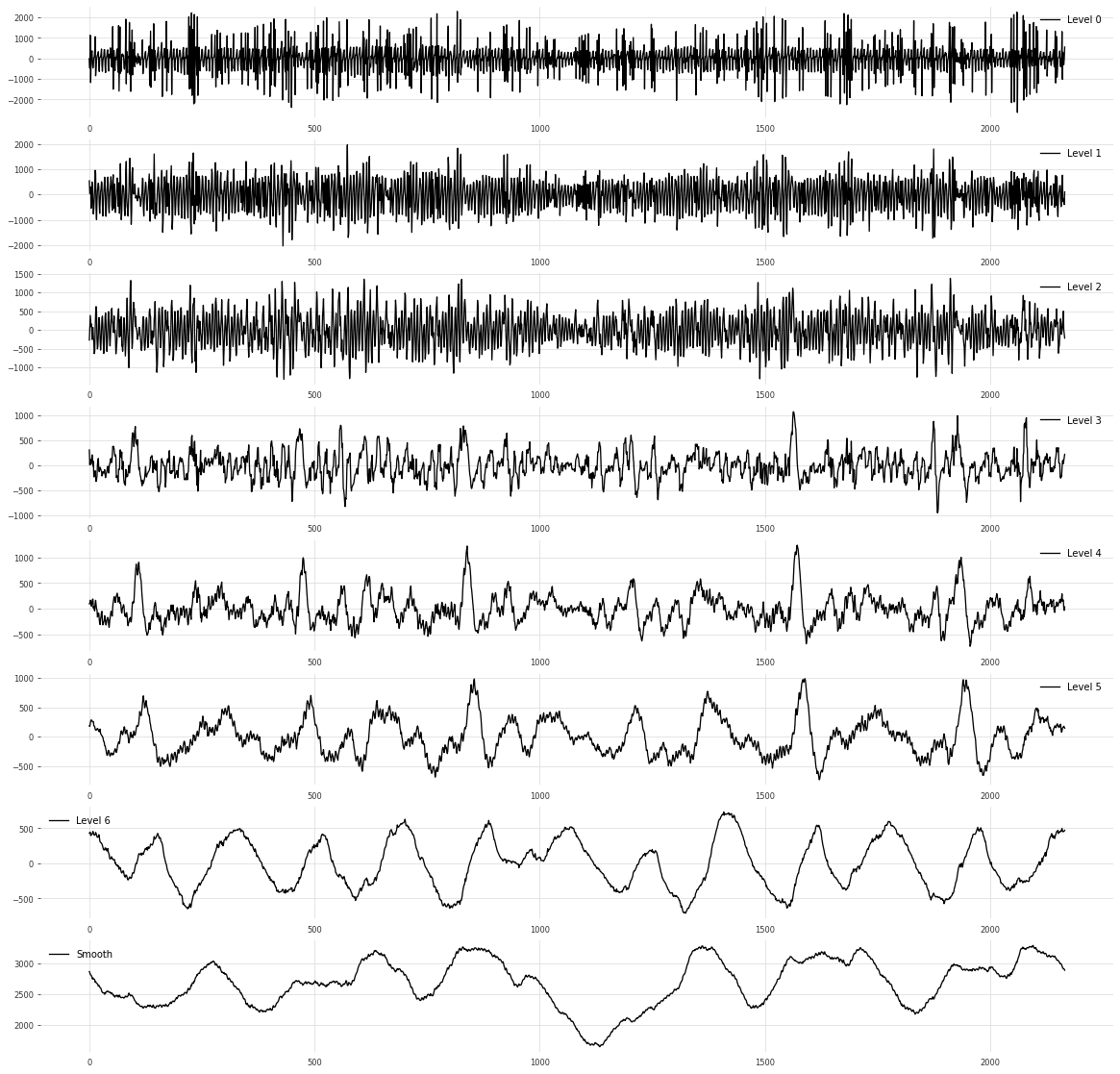}
    \caption{MODWT decomposition of the website traffic dataset. The first curve represents the original time series; the last represents the smooth series, i.e., the scaling coefficient of the time series after computing the MODWT algorithm. The other seven curves (in between) represent the different level coefficient calculated by the MODWT algorithm. The filter used for this decomposition is the Haar filter.}
    \label{waveletsdecompostion}
\end{figure}

In our proposed approach, we use the MODWT algorithm (discussed in Section \ref{modwt}) as a data pre-processing method to capture more explicit information about the observed time series. The MODWT algorithm will provide us with $J$ detail coefficients (higher frequency bands) and a smooth (trend component) of the available time series $Y_t \; (t=1,2,\ldots,N)$, where $N$ is the number of past observations. We show a pictorial view of the MODWT decomposition using a Haar filter for the publicly available website traffic data in Fig. \ref{waveletsdecompostion}. Wavelet levels are chosen as an integer value that specifies the number of time series decompositions into smooth details. We set it as $J+1 = \lfloor \log_{e}{N} \rfloor$ based on the recommendation in \cite{percival2000wavelet}. These high and low-frequency decomposed series are termed wavelet (details), and scaling (smooth) coefficients that can track the original series as:
$$Y_t = \sum_{j=1}^J D_{j,t} + S_{j,t},$$
where $D_{j,t}$ is the $j^{th}$ level details and $S_{j,t}$ is the smooth of the decomposed time series. The MRA-based MODWT decomposed series can handle nonstationarity and seasonality in the time series datasets and capture the overall trend of the series.
Each level of the time series and the smooth component are then passed into several independent Transformers for the prediction task. Time series datasets are passed through two parallel networks starting both an embedding layer. The encoder embedding takes the input of the batch train data and the decoder takes the output of the batch train data. Further, the output of both the encoder and decoder will enter into a second decoder. The encoder is composed of multi-head attention, a normalization layer, and a feedforward layer. The first decoder has a masked multi-head attention layer and normalization, and the second decoder has the same structure as the encoder. We apply a linear and a softmax layer to the output of the second decoder, which gives us the weight parameters of our model. Once all the epochs are done, we have our final model ready for making predictions and one-step ahead forecasts. In the end, we predict a time series for each level of decomposition and the smooth size of the forecasting horizon with their respective transformer model and reverse the process of MODWT with inverse MODWT as discussed in Sec. \ref{modwt}. The result of the inverse MODWT will be our final prediction that could be compared with the test time series. Multi-step ahead forecasts are generated using an iterative method applied to the fitted W-Transformers model. To generate $h$-step ahead forecasts $(\hat{Y}_{N+h})$, we use the following equation:
\begin{equation*}
    \hat{Y}_{N+h} = \sum_{j=1}^J \hat{D}_{j,N+h} + \hat{S}_{J,N+h},
\end{equation*}
where
\begin{align*}
 \hat{D}_{j,N+h} & = f(D_{j,1},D_{j,2},\ldots,D_{j,N}) ; \; j= 1,2, \ldots, J, \\
 \hat{S}_{J,N+h} & = f(S_{J,1}, S_{J,2}, \ldots, S_{J,N}),
\end{align*}
and $f$ is the Transformer model. 
We describe the mathematical formulation of the Transformer model that we use in our ensemble W-Transformers framework to keep the current work self-contained, following \cite{vaswani2017attention}. Once the nonstationarity and seasonality are handled by the MODWT algorithm, then Transformers model the long-range dependencies and interactions in sequential decomposed datasets. Mathematically, the self-attention mechanism of Transformers is defined as follows: 
$$Attn(Q,K,V) = \text{softmax}\left(\frac{Q K^{T}}{\sqrt{d}}\right) V,$$
where $Q$ is the query matrix, $K$ is the key matrix, $V$ is the value matrix, and $d$ is the dimension of $Q$, $K$, and $V$. Next, the softmax function is used to obtain the weights on the values to be put into a matrix $Q$. Now, the keys, queries, and values are then linearly projected, and these projections perform attention function in parallel yielding the output values. Theoretically, we define these heads as follows:
$$head_{p} = Attn(QW^{Q}_{p},KW^{K}_{p},VW^{V}_{p}),$$
where the projections are parameter matrices, denoted by $W^{Q}_{p}, W^{K}_{p}, W^{V}_{p}$. Finally, the multi-head attention is simply a concatenation of the $head$ for $m$-times, where $m$ is an hyper-parameter that is usually set to a pre-specified integer value $8$ (as in \cite{vaswani2017attention}) and $d$ the dimension of $Q$ and $V$ defined by $d = \frac{d_{model}}{m}$, where $d_{model}$ is the dimension of the output of the embedding layer. To jointly attend the information from various subspaces, we use multi-head attention defined as follows:
$$\text{MultiHead}(Q,K,V) = \text{Concat}(head_{1},...,head_{m})W^{O},$$
where $W^{O}$ is the projection matrix of the output layer.

\begin{figure*}
\centering
\includegraphics[width=0.75\linewidth]{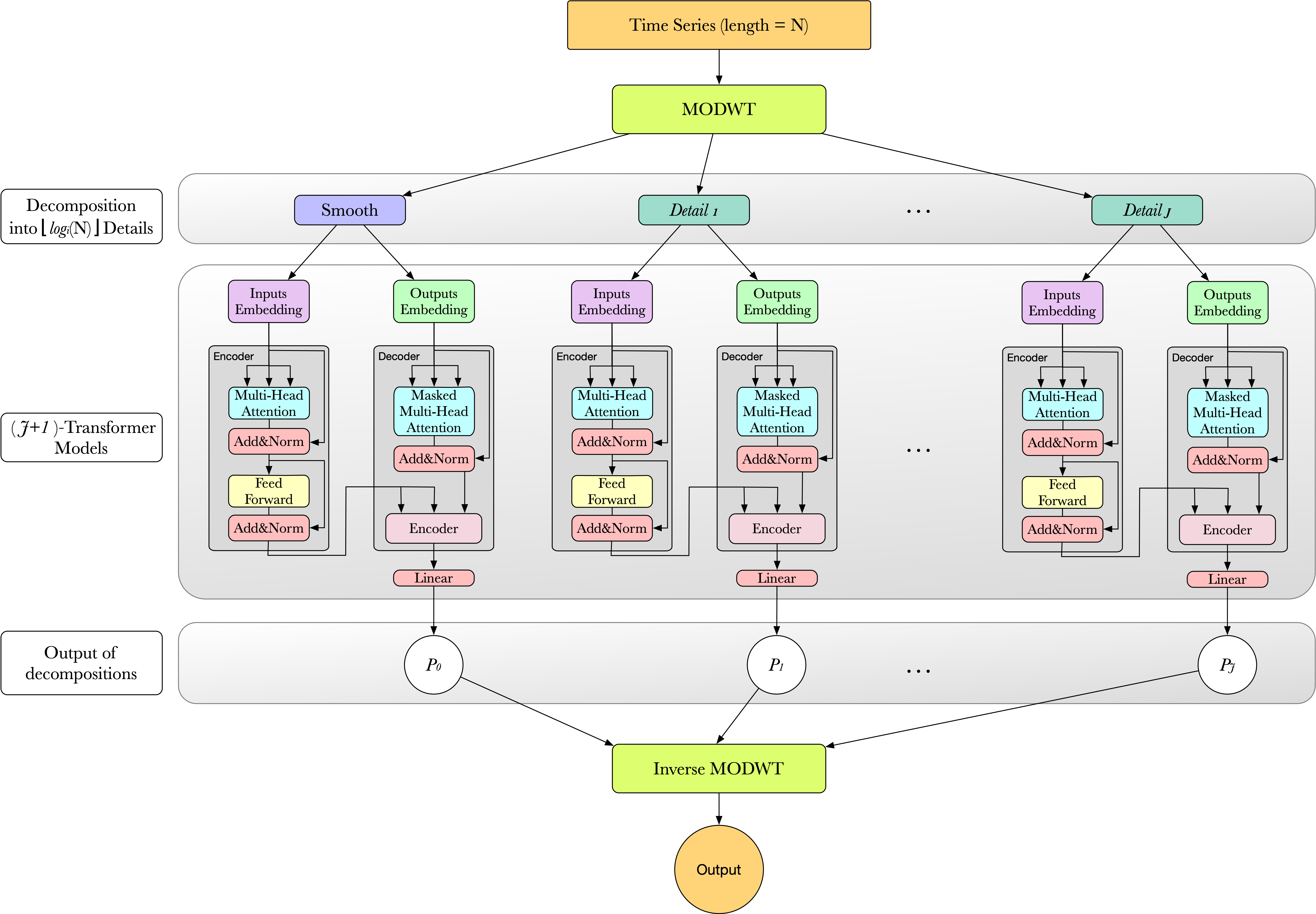}
\caption{Diagram representing the proposed W-Transformers architecture; $P_i \; (i:0,...,j)$ represent the prediction.}
\label{fig_sim}
\end{figure*}

\begin{algorithm}
  \caption{W-Transformers}\label{W-transformers}
  \begin{algorithmic}
  \Require{The original Time Series (training set): $Y_t$ \; (t=1,2,\ldots,N)} 
  \Ensure{The prediction of the testing series: $\hat{Y}_{N+h}$}
      \State $NForecast \gets \text{Number of data point of testing set} \; (h);$
      \State $J+1 \; \text{-decomposition} \gets \text{MODWT}(Y_t);$
      \For{\texttt{$i$ in $J+1$}}
        \State $train \gets (J+1)$-$Levels[i][1:N];$
        \State $test \gets (J+1)$-$Levels[i][N+1:N+h];$
        \For{\texttt{each $epochs$}}
            \State $embedIn \gets \text{Embedding1}(train);$
            \State $embedOut \gets \text{Embedding2}(train);$
            \State $encodOut \gets \text{Encoder}(embedIn);$
            \State $decod1Out \gets \text{Decoder1}(embedOut);$
            \State $decod2Out \gets \text{Decoder2}(encodOut,decod1Out);$
            \State $Model \gets softmax(Linear(decod2Out));$
        \EndFor
        \State $transformerResults[i] \gets prediction(Model,test);$
      \EndFor
      \State $\hat{Y}_{N+h} \gets $InverseMODWT$(TransformerResult);$
  \end{algorithmic}
\end{algorithm}
A detailed explanation of the W-Transformers is given in Algorithm \ref{W-transformers} and a schematic diagram is also presented in Fig. \ref{fig_sim}. 

\section{Experimental Analysis}\label{Experiment}
In this section, we present the used datasets, the performance measures, the detailed description of the implementation scheme of the proposed W-Transformers, and the baseline methods for comparisons. 
\subsection{Data}
We considered seven publicly available datasets (see Table \ref{tab:Data}) from various applied fields for our experiments. For example, one dataset is from the stock market, three are from epidemiological fields, two are related to internet traffic, and the last one is about the sunspot data.
\begin{itemize}
    \item \textbf{Netflix}: This dataset is the Netflix closing stock price from June 2021 to June 2022. The data are coming from Yahoo Finance and are daily data.
    \item \textbf{Website Traffic}: This dataset is the number of first-time visitors to a website. This dataset is publicly available on Kaggle and contains five years of daily time points.
    \item \textbf{Sunspot}: This standard time series dataset contains daily total sunspot number derived by the formula: $R= Ns + 10Ng$, with $Ns$ the number of spots and $Ng$ the number of groups counted over the entire solar disk \cite{godahewa2021monash}. 
    \item \textbf{Colombia Dengue}: This dataset represents the weekly number of infected people with dengue in Colombia \cite{chakraborty2019forecasting}. 
    \item \textbf{Japan Flu}: This dataset reports weekly influenza-like cases in Japan \cite{panja2022epicasting}.
    \item \textbf{Bangkok Dengue}: This dataset is the monthly number of infected people with Dengue in Bangkok \cite{chakraborty2019forecasting}.
    \item \textbf{Network Analytics}: This dataset represents the percentage of network outbound utilization every 5-min and is available on Kaggle.
\end{itemize}
On these seven univariate datasets, we applied KPSS Test for stationarity, Terasvirta test for linearity, and Hurst exponent check for long-term dependencies \cite{chakraborty2022nowcasting, hyndman2018forecasting}. Results of these tests on different characteristics are shown in Table \ref{tab:Data}. Other than the Japan Flu dataset, all other datasets exhibit strong nonstationarity and long-term dependency. Only NFLX and Sunspot datasets show linear structure in the time series. 
\begin{table*}
\scriptsize
    \begin{center}
      \caption{Data Characteristics of seven time series datasets}
      \label{tab:Data}
      \renewcommand{\arraystretch}{2}
      \begin{tabular}{|l|c|c|c|c|c|c|c|} 
      \hline
        \textbf{Data Name} & \textbf{Frequency} & \textbf{Observations} & \textbf{Test Size} & \textbf{Min-Max Value} & \textbf{Behaviour} \\
        \hline
          NFLX & Daily & 254 & 30/90 & 166.37 - 691.69 & Non-stationary, Linear, Long-term dependant\\
        \hline
          Website Traffic & Daily & 2167 & 30/90 & 2 - 5541 & Non-stationary, Non-linear, Long-term dependant\\
        \hline
          Sunspot & Daily & 2663 & 30/90 & 0 - 172 & Non-stationary, Linear, Long-term dependant\\ 
        \hline
          Colombia Dengue & Weekly & 626 & 26/52 & 164 - 3354 & Non-stationary, Non-linear, Long-term dependant\\
        \hline
          Japan Flu & Weekly & 964 & 26/52 & 0 - 1656 & Stationary, Non-linear, Long-term dependant\\
        \hline
          Bangkok Dengue & Monthly & 180 & 12/24 & 29 - 8337 & Non-stationary, Non-linear, Long-term dependant\\
        \hline
          Network Analytics & 5-min & 25631 & 576/1152 & 4.82e-05 - 99.9955 & Non-stationary, Non-linear, Long-term dependant\\
        \hline
      \end{tabular}
      \renewcommand{\arraystretch}{1}
    \end{center}
  \end{table*}

\subsection{Performance Measures}
To comprehensive analysis and evaluation of different methods, we considered four commonly used performance metrics including Root Mean Squared Error (RMSE), Mean Absolute Error (MAE), Symmetric Mean Absolute Percentage Error (sMAPE), and Mean Absolute Scaled Error (MASE) \cite{hyndman2018forecasting}. The lower values of these measures indicate better forecasting results. 

\subsection{Protocol}
To perform MODWT and inverse MODWT, we used the python package 'Pywavelet' \cite{Lee2019PyWavelet}. To use MODWT for time series analysis, we carefully chose the number of levels for our decomposition to get the optimal performance in the wavelet domain as discussed in Sec. \ref{proposed}. Once our proposed model is implemented, we computed the different metrics for all the baseline models for different forecasting horizons on each dataset. The different forecasting horizons are defined as short-term forecasting and long-term forecasting, as shown in the test size column in Table \ref{tab:Data}. The implementation of Transformers on the decomposed datasets is done using `darts' \cite{Herzen2021Darts} Python implementation tools by setting the number of epochs to be 200 for all the series. We tested several configurations for the input/output length, and regarding the result, we normalized the value to 12/1 for every dataset. The number of heads $m$ is set to $8$ and the dimension of the model, $d$, is 16, and the batch size is set to 32. We have used two encoder layers as well as two decoder layers. The dropout is set to 0.1, and the activation used in the W-Transformers model is `relu' \cite{zerveas2021transformer}.
Our proposed model was trained on GPU (graphics processing unit) but can also be computed using a CPU (central processing unit). 

\subsection{Results}
\begin{figure}[!t]
    \includegraphics[width=0.6\linewidth]{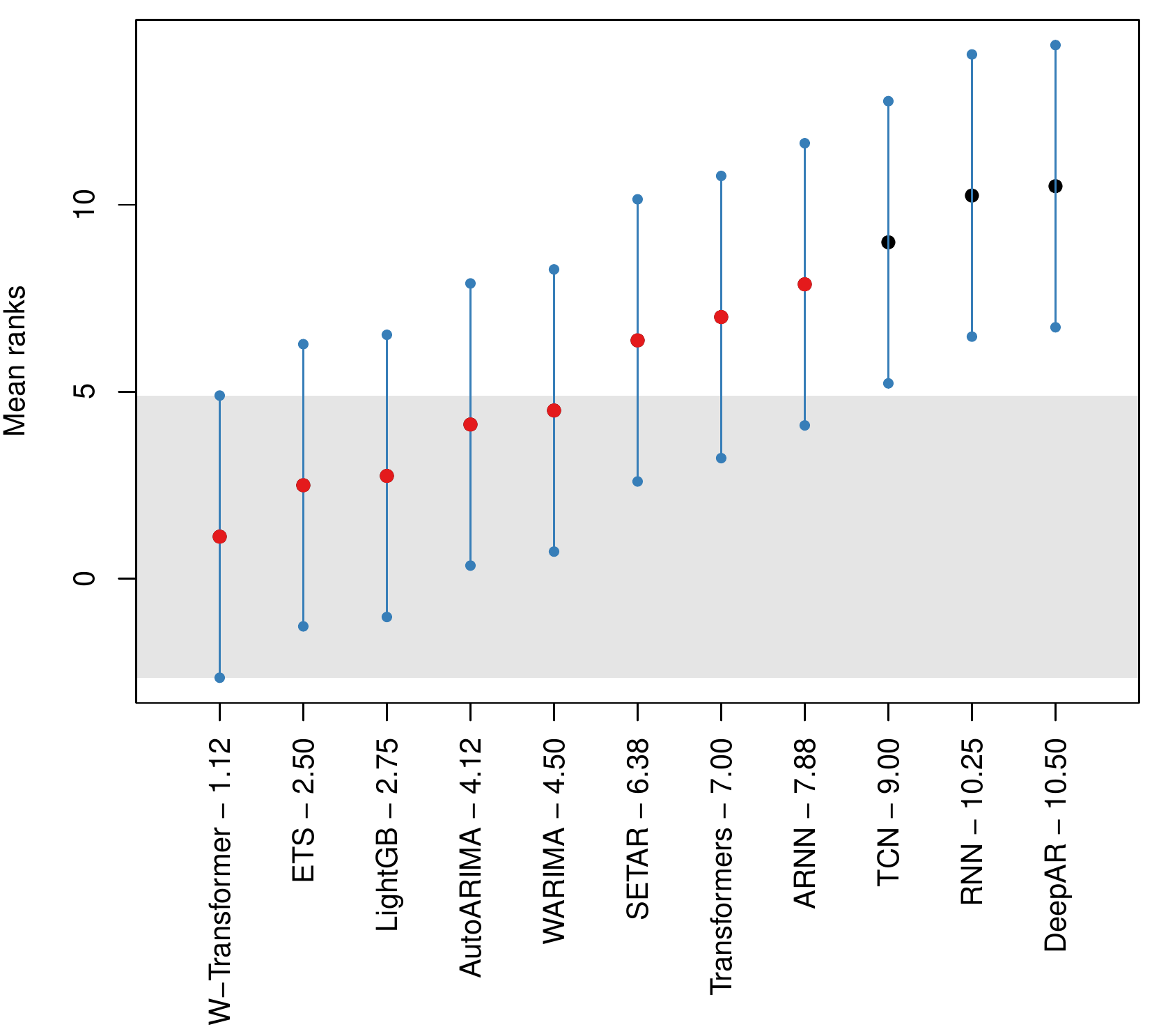}
    \caption{Average model ranking regarding the four evaluation metrics for short-term forecasting using multiple comparisons with best(MCB). The number after the model name represents the average ranking.}
    \label{Rank_short}
\end{figure}

\begin{figure}[!t]
    \includegraphics[width=0.6\linewidth]{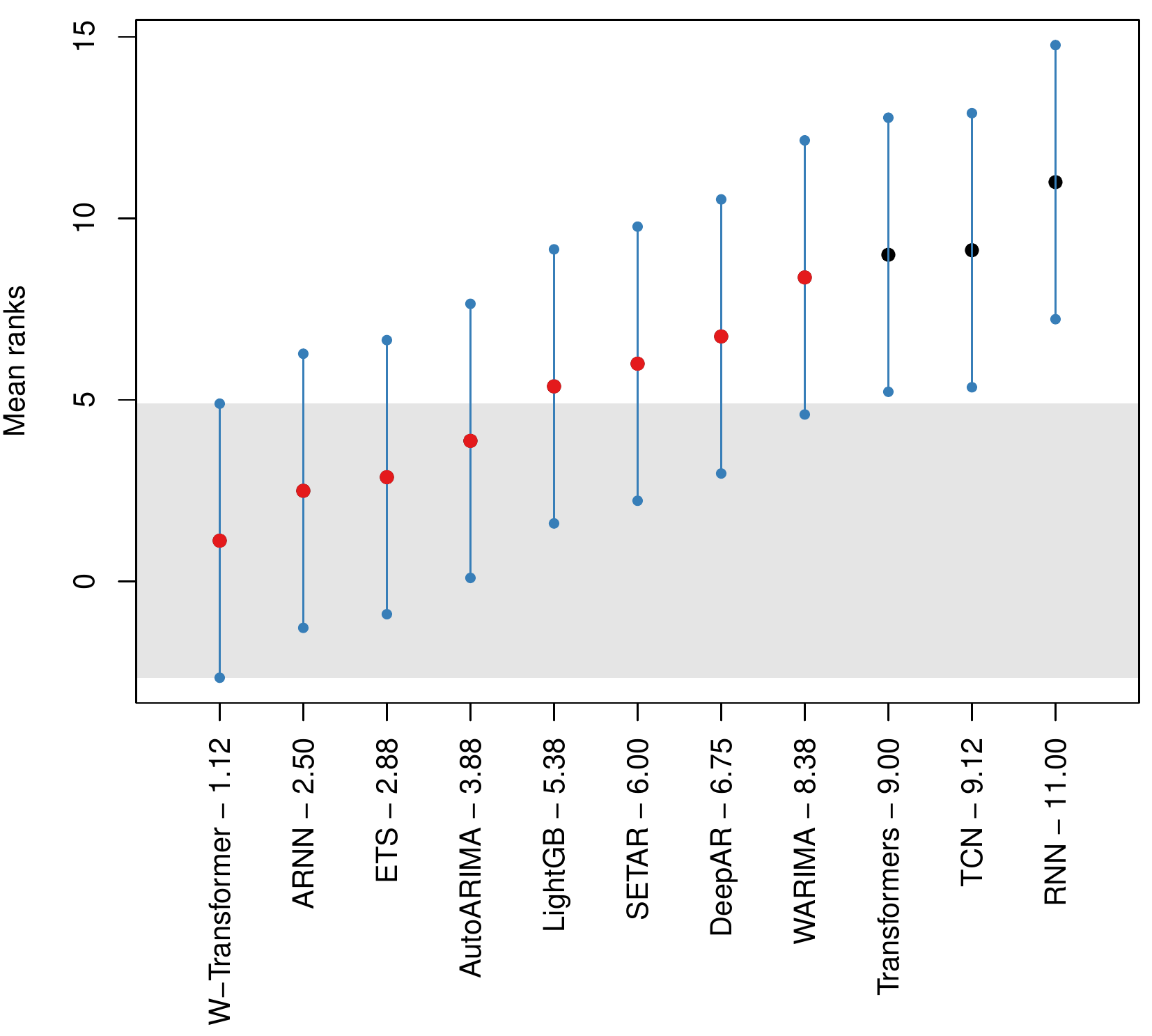}
    \caption{Average model ranking regarding the four evaluation metrics for long-term forecasting using multiple comparisons with best (MCB). The number after the model name represents the average ranking.}
    \label{Rank_Long}
\end{figure}

For a comprehensive comparison, we computed the results of our proposed W-Transformers and ten baseline models on seven datasets with two different forecast horizons. The experimental results for short-term and long-term forecasting are shown in Table \ref{tab:Short} and Table \ref{tab:Long}, respectively. The experiments show that the performance of the models mainly depends on the length of the forecast horizons. W-Transformers is, in fact, more efficient for long-term forecasting. This comes from the MODWT decomposition; wavelets can handle nonstationarity and Transformers work mostly well for long-term forecasting. To demonstrate the statistical robustness of the obtained results, we applied multiple comparisons with the best (MCB) test by combining all the results based on four performance metrics and reporting the results in Figures \ref{Rank_short} and \ref{Rank_Long}. The overall performance of our W-Transformers is superior compared to all deep learning models and statistical forecasters considered in this study. For small sample-sized datasets, the performance of statistical methods is comparatively better than that of  W-Transformers. However, the proposal performed better than all the baseline models for large temporal datasets (for example, the Network Analytics dataset, etc.). All the data and codes are available at \url{https://github.com/CapWidow/W-Transformer} for public use to support the principle of reproducible research.

\begin{table*}
\scriptsize
\centering \caption{Short-Term Comparison of proposed W-Transformers (W-Trans.) with statistical, machine learning, and deep learning models in terms of RMSE, MAE, sMAPE, and MASE ((Best results in \textbf{Bold} and second best in \emph{italic}).}
    \begin{tabular}{|c|c|c|c|c|c|c|c|c|c|c|c|c|}
        \hline
        Data & Metrics & Auto & WARIMA & ETS & SETAR & LightGBM & ARNN & RNN &  TCN & DeepAR & Transformer  & {\color{blue}{W-Trans.}} \\
        &    & ARIMA \cite{hyndman2008automatic} & \cite{aminghafari2007forecasting} & \cite{hyndman2008forecasting} &  \cite{tong1990non} & \cite{ke2017lightgbm} & \cite{faraway1998time}  &   \cite{hochreiter1997long} & \cite{chen2020probabilistic} & \cite{salinas2020deepar}  &  \cite{vaswani2017attention} & {\color{blue}{(Proposed)}} \\ \hline
{NFLX} 
        & RMSE & 89.06 & 87.70 & \textbf{13.09} & 116.54 & 81.31 &153.90 &
        172.62 & 90.02 & 149.27 & 203.54 & \emph{45.71} \\ 
        & MAE & 75.12	& 77.15	& \textbf{10.21} &	95.24 &79.75 & 148.82&	172.34 &	89.24 &	148.95 	& 202.66 &	\emph{40.00}\\
        Stock & sMAPE &  54.41	& 54.77 &	\textbf{5.37} &	78.63 & 34.77 & 136.82 &	166.41 & 38.13 &	129.14 &	198.97 &	\emph{24.26}\\
        & MASE & 8.32 & 12.00 &	\textbf{1.59} &	14.81 & 8.84 &	23.15 & 19.10 
        &	9.89 &	16.51 &	22.46 & \emph{4.43}\\ \hline
{Website}  
        & RMSE  & 812.65 & 1110.80 &	849.02 &	\emph{786.99} & 899.36 & 989.77& 2227.65 &	869.51 & 1628.14 &	1584.98 &	\textbf{728.70}\\ 
        & MAE &\textbf{435.80} & 762.14 &	579.08 &	633.82 &       786.15 & 877.43&  2080.15 & 542.88 & 1536.25 & 1257.2 &	\emph{473.82}\\
        Traffic & sMAPE &\textbf{25.32} &  36.05 &	31.11 &	35.05 & 45.36 & 55.25& 158.91 &	30.95 & 98.13 &	79.10 & \emph{27.35}\\
        & MASE  &\textbf{0.48} &  0.97 & 0.73 & 0.80 & 0.87 &1.11 & 2.29 &	0.60 & 1.69 & 1.38 & \emph{0.52}\\ \hline
{Sunspot} 
        & RMSE  & 51.97 & 50.72 &	73.25 &	85.53 &	84.32 & \emph{39.22}&  82.35 &	4448.45 &	79.30 &	59.37 &	\textbf{34.05}\\
        & MAE & 42.46 & 40.74 &	62.87 &	76.05 &	75.86 & \emph{33.55} & 70.90 &	2832.47 &	70.61 &	48.72 &	\textbf{30.82}\\
        & sMAPE & 47.79 & 45.37 & 84.16 & 116.78 & 119.15 & \emph{37.80} &	 104.36 &	179.82 & 111.39 &	57.52 &	\textbf{37.23}\\
        & MASE & 7.13 & \emph{2.91} &	4.49 &	5.43 & 12.73 & \textbf{2.40} & 11.90 &	475.42 & 11.85 &	8.18 &	5.17\\ \hline
{Colombia}
        & RMSE  & 949.82 & 840.50 & 917.22 & 809.30 & 702.48 & 1140.47& 
        1263.69 & 1283.29 & 1153.45 & \emph{635.18} & \textbf{634.41} \\
        & MAE & 877.08 & 793.21 & 843.16 & 741.10 & 642.27 & 1058.74& 
        1207.63 & 1224.61 & 1091.73 & \textbf{537.16} & \emph{541.77}\\
        Dengue & sMAPE & 54.79 & 51.26 & 53.34 & \emph{48.81} &   \textbf{44.10} &61.92 & 184.26 & 183.90 & 150.59 & 50.21 & 51.03\\
        & MASE  & 9.07 & 8.20 & 8.72 & 7.66 & 7.40 & 10.95& 13.92 & 14.12 &   12.59 & \textbf{6.19} & \emph{6.25}\\\hline
{Japan}
        & RMSE  & 114.51 & 114.87 & \textbf{6.84} & 239.40 & 29.78 & 283.89&  \emph{10.71} & 11.09 & 151.34 & 20.58 & 94.14\\
        & MAE & 108.39 & 103.27 & \textbf{5.69} & 210.15 & 13.43 & 228.04& 
        8.06 & \emph{7.53} & 114.48 & 18.93 & 90.51\\
        Flu & sMAPE & 171.98 & 167.0 & \emph{73.38} & 176.98 & \textbf{66.38} & 163.75& 169.00 & 90.78 & 157.19 & 189.51 & 170.81 \\
        & MASE & 2.12 & 18.18 & 1.00 & 37.00 & 0.26 & 40.15& 
        \emph{0.16} & \textbf{0.15} & 2.24 & 0.37 & 1.77\\\hline
{Bangkok}
        & RMSE &  493.09 & 779.49 & 424.68 & \textbf{415.42} & \emph{424.15} &  518.08& 886.48 & 998.86 & 841.74 & 827.06 & 815.52\\
        & MAE & 441.83 & 743.68 & \emph{362.42} & 375.64 &          \textbf{359.32} & 457.00& 754.74 & 917.99 & 701.68 & 684.24 & 681.20\\
        Dengue & sMAPE & 65.43 & 76.93 & \emph{53.95} & 56.23 &          \textbf{53.13} &  67.98& 188.83 & 189.04 & 152.32 & 142.41 & 150.26\\
        & MASE & 1.63 & 3.26 & \emph{1.59} & 1.65 & \textbf{1.33} & 2.00&    2.79 & 3.39 & 2.59 & 2.53 & 2.52\\\hline
{Network}
        & RMSE &  25.8 & \emph{25.08} & 52.43 & 44.63 & 26.76 & 32.57& 
        31.68 & 1.28$E^{95}$ & 53.70 & 37.58 & \textbf{12.09} \\
        & MAE & 23.02 & 19.69 & 48.49 & 41.10 & \emph{18.13} & 29.73&
        28.98 & 7.76$E^{93}$ & 49.73 &  34.39 & \textbf{10.91}\\
        Analytics & sMAPE & 88.35 & 83.29 & 114.08 & 108.25 & \emph{77.20} & 97.14& 96.13 & 199.76 & 114.96 & 101.79 & \textbf{60.09} \\
        & MASE &  3.94& 3.43 & 8.45 & 7.16 & \emph{3.08} & 5.18& 8.85&   1.32$E^93$ & 8.45 &   5.98&   \textbf{1.72}   \\ \hline
    \end{tabular}
    \label{tab:Short}
\end{table*}

\begin{table*}
\scriptsize
\centering \caption{Long-Term Comparison of proposed W-Transformers (W-Trans.) with statistical, machine learning, and deep learning models in terms of RMSE, MAE, sMAPE, and MASE ((Best results in \textbf{Bold} and second best in \emph{italic}).}
    \begin{tabular}{|c|c|c|c|c|c|c|c|c|c|c|c|c|} 
    \hline
        Data & Metrics & Auto & WARIMA & ETS & SETAR & LightGBM & ARNN & RNN &  TCN & DeepAR & Transformer  & {\color{blue}{W-Trans.}} \\
        &    & ARIMA \cite{hyndman2008automatic} & \cite{aminghafari2007forecasting} & \cite{hyndman2008forecasting} &  \cite{tong1990non} & \cite{ke2017lightgbm} & \cite{faraway1998time}  &   \cite{hochreiter1997long} & \cite{chen2020probabilistic} & \cite{salinas2020deepar}  &  \cite{vaswani2017attention} & {\color{blue}{(Proposed)}} \\ \hline
{NFLX}
        & RMSE &  269.52& 696.59& 276.92&       270.18&  \emph{210.78} & \textbf{160.85}& 311.52& 254.39& 300.90& 328.69& 237.67\\
        & MAE &   245.91& 615.44& 252.77&       247.12&   \emph{189.15}& \textbf{125.40} &297.47& 241.01& 286.36& 312.38& 219.75\\
        Stock & sMAPE & 127.19& 165.81& 128.90&       60.33& \emph{50.70}& \textbf{37.80} &187.98& 149.49&        173.50& 198.81& 107.29\\
        &  MASE & 30.93&  62.97&  25.86& 25.28& \emph{23.79}& \textbf{12.83}& 37.41&  30.31&         36.02&  39.29&  27.64\\
      \hline
{Website}
        & RMSE &  \emph{959.73}&   1281.64& 1192.66& 1082.51& 1399.02& 1356.29&2593.36& 1166.51&  2010.79& 2638.05&  \textbf{847.41}\\
        & MAE  &  \textbf{624.22}& 975.38&  864.14&  921.82&  1261.93& 1065.48 &2413.45& 782.65&   1875.34& 2470.93&  \emph{634.74}\\
        Traffic & sMAPE & \textbf{29.49}&  39.48&   36.31&   43.89&   64.28& 41.23&164.07&  33.91&    107.14&  180.14&   \emph{31.02}\\
        & MASE &  \textbf{0.69}&   1.10&    0.98&    1.04&    1.39& 1.21&2.66&    0.86&     2.07&    2.73&     \emph{0.70}\\
      \hline
{Sunspot}
        &  RMSE &  44.61& 41.48&       \emph{37.46}&  57.06& 66.99&71.83&74.16&  41900.15& 52.50& 40.63& \textbf{30.07}\\
        &  MAE &   34.93& 33.05&       \emph{30.72}&  45.67& 53.09&56.93&63.75&  15410.47& 41.78& 32.36& \textbf{22.63}\\
        &  sMAPE & 44.15& 41.48&       \emph{38.21}&  62.91& 84.14&97.60&108.69& 166.08&   65.21& 40.40& \textbf{30.09}\\
        &  MASE &  5.98&  \emph{2.80}& \textbf{2.60}& 3.87&  9.08&4.82&10.91&  2636.93&  7.15&  5.54&  3.87\\
      \hline
{Colombia}
        &  RMSE &   998.32 &  863.05 &  997.25 &  \textbf{707.28} &  1183.99&\emph{839.24}& 2114.89 &  2126.41 &  2016.42&  1549.02  &  1452.04\\
        &  MAE &    802.61 &  736.40 &  802.10 &  \textbf{640.13} &  946.09&\emph{683.25}& 1949.85 &  1966.53 &  1842.58&  1324.07  &  1332.04\\
        Dengue&  sMAPE &  40.99 &   38.92 &   40.97 &   \textbf{35.15} &   45.43&\emph{36.62}& 188.75 &   193.33 &   165.68 &  88.37  &    98.81\\
        &  MASE &   4.86 &    4.46&    4.86 &    \textbf{3.88} &    11.83&\emph{4.14} & 24.38 &    24.59 &    23.04 &   16.55 &     16.65\\

      \hline
{Japan}
        &  RMSE &  \emph{164.41} &  196.65 &  186.15 &  297.30 & 192.54&239.31& 171.51 &        194.94 &        179.61 & 326.55  & \textbf{76.21}\\
        & MAE &    145.25 &         174.17 &  171.63 &  281.93 & 126.51&199.93& 114.01 &        \emph{106.89} & 163.67 & 276.56  & \textbf{58.98}\\
        Flu & sMAPE &  130.81 &         136.76 &  134.94 &  142.31 & \emph{106.52}&126.77& 130.00 & 131.07 &        133.18 & 131.81  & \textbf{103.19}\\
        & MASE &   2.89 &           4.83 &    3.95 &    6.49 &   2.52& 4.60 &2.27 &          \emph{2.13} &   3.26 &   5.51  &   \textbf{1.17}\\
        
      \hline
{Bangkok}
        & RMSE &  4152.04 & 1889.92 &       3454.05 &  2153.80 &          1492.16&819.90& 824.70 & 1032.96 & 786.21 &      \emph{767.52} & \textbf{735.00}\\
        & MAE &   4131.17 & 1756.66 &       3423.33 &  1486.24 &          1418.78&678.36& 681.73 & 877.07 &  634.59 &      \emph{611.18} & \textbf{608.30}\\
        Dengue & sMAPE & 152.16 & 119.20 & 145.50 &   114.83 &   \emph{108.60} &\textbf{76.91}& 187.26 & 157.79 &  151.00 &      136.43 &            154.62\\
        & MASE &  15.51 &   7.57 &          14.75 &    6.40 &             5.33&2.92& 2.56 &   3.29 &    2.38 &        \emph{2.29} &   \textbf{2.28}\\
        
      \hline
{Network}
        & RMSE &   27.26 &  43.94 &  23.65 &         40.58 &  28.56& 24.71 &43.00 &     $\infty$ &     \emph{22.51} &  29.21 &  \textbf{19.00}\\
        & MAE &    24.08 &  39.06 &  \emph{18.31} &  35.97 &  25.19& 21.99&37.98 &  9.66$ E^{182} $ &   19.09 &        25.80 &   \textbf{15.96}\\
        Analytics & sMAPE &  78.28 &  94.56 &  \emph{70.46} &  91.69 &  80.02& 75.80&93.34 &  199.80 &       71.52 &        80.64 &  \textbf{60.31} \\
        & MASE &   4.10 &    6.49 &   \emph{3.04} &   5.97&    4.29& 3.66&6.46 &   1.64$ E^{182} $ &   3.25 &         4.39 &   \textbf{2.71} \\ \hline
          
    \end{tabular}
      \label{tab:Long}
\end{table*}

\begin{figure*}[!t]
\centering
    \includegraphics[width=0.6\linewidth]{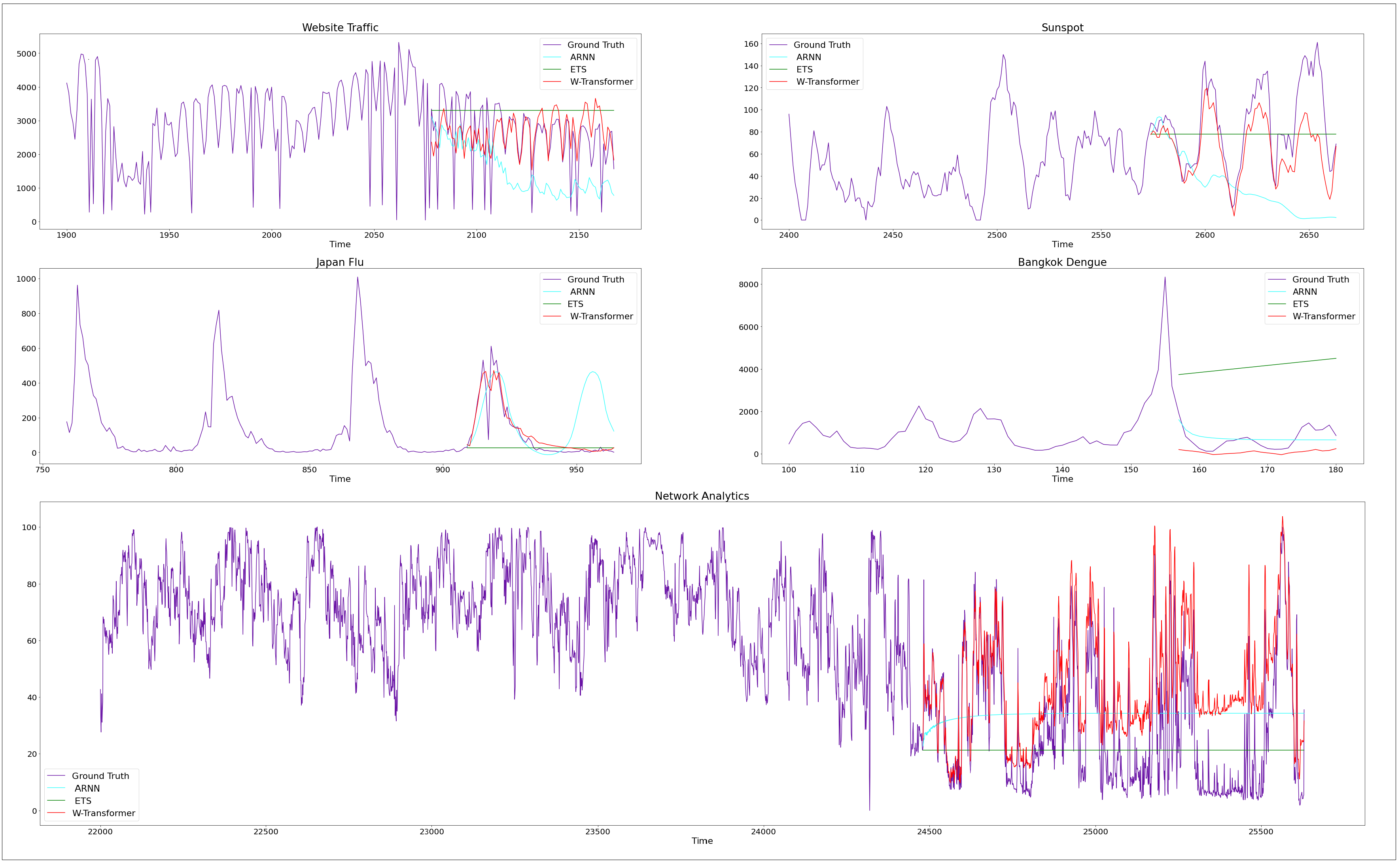}
    \caption{The predicted curve of W-Transformers (red), ETS (green), and ARNN (blue) were superposed over the test data (violet) for long-term forecasting.}
\end{figure*}

\section{Conclusions and Discussions}
We proposed a novel approach called W-Transformers for time series forecasting. Our model uses wavelet decomposition within the Transformers framework. The proposed model achieves much better results on average than several existing models. Moreover, our proposed approach consistently outperforms all conventional Transformers-like methods on the considered datasets. It is worth noting, nevertheless, that our proposed approach works better with large-size datasets. Statistical models can easily outperform deep learning-based approaches with too small data sizes. Another important finding from W-Transformer compared with the benchmarks is that the proposal is superior for long-term forecasting. In future work, extending the proposed method to work better with small data sizes will be of interest. Improving the results for short-term forecasting is also a key challenge and can be considered a future scope of the study. Applying the proposed W-Transformers approach to multivariate time series is under investigation. 

 \section*{Acknowledgment}
 
 The support of TotalEnergies is fully acknowledged. Lena Sasal (PhD Student) and Abdenour Hadid  (Professor,  Industry Chair at SCAI Center of Abu Dhabi) are funded by TotalEnergies collaboration agreement with Sorbonne University Abu Dhabi.


\bibliographystyle{IEEEtran}
\bibliography{ref}


\end{document}